# Large Margin Structured Convolution Operator for Thermal Infrared Object Tracking

Peng Gao, Yipeng Ma, Ke Song, Chao Li, Fei Wang, Liyi Xiao
Shenzhen Graduate School, Harbin Institute of Technology, China

***Abstract*—Compared with visible object tracking, thermal infrared (TIR) object tracking can track an arbitrary target in total darkness since it cannot be influenced by illumination variations. However, there are many unwanted attributes that constrain the potentials of TIR tracking, such as the absence of visual color patterns and low resolutions. Recently, structured output support vector machine (SOSVM) and discriminative correlation filter (DCF) have been successfully applied to visible object tracking, respectively. Motivated by these, in this paper, we propose a large margin structured convolution operator (LMSCO) to achieve efficient TIR object tracking. To improve the tracking performance, we employ the spatial regularization and implicit interpolation to obtain continuous deep feature maps, including deep appearance features and deep motion features, of the TIR targets. Finally, a collaborative optimization strategy is exploited to significantly update the operators. Our approach not only inherits the advantage of the strong discriminative capability of SOSVM but also achieves accurate and robust tracking with higher-dimensional features and more dense samples. To the best of our knowledge, we are the first to incorporate the advantages of DCF and SOSVM for TIR object tracking. Comprehensive evaluations on two thermal infrared tracking benchmarks, i.e. VOT-TIR2015 and VOT-TIR2016, clearly demonstrate that our LMSCO tracker achieves impressive results and outperforms most state-of-the-art trackers in terms of accuracy and robustness with sufficient frame rate.**

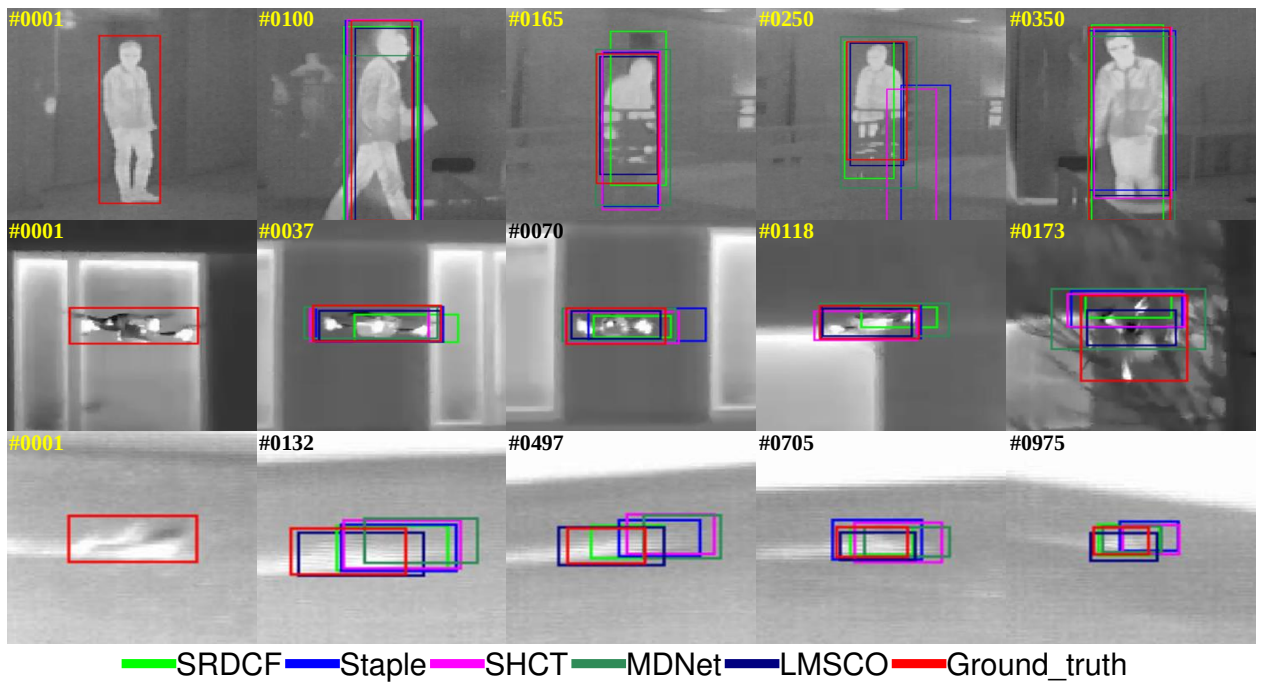

Fig. 1: A comparison of our method LMSCO (in Navy-Blue) with four state-of-the-art TIR trackers SRDCF [6] (in Green), Staple [14] (in Blue), SHCT [15] (in Magenta) and MDNet [16] (in SeaGreen) on three thermal infrared sequences *hiding*, *quadrocopter* and *ragged* in VOT-TIR2016 [17], respectively. Besides, we crop the target patches to enhance the visibility and give the ground-truth of each frame (in Red). Best viewed in color.

## I. Introduction

Visible object tracking is one of the most challenging problems in computer vision. It has been applied to various applications, such as video surveillance, service robots, UAV monitoring, human-computer interaction and so on. In the past decade, despite a particularly large number of tracking algorithms [1]–[7] have been proposed and show excellent performances for visual object tracking, there still remain numerous unsolved challenges such as illumination variations and deformations. In contrast to visible object tracking, thermal infrared (TIR) object tracking is not sensitive to the variations of illumination, and it can track the target in total darkness [8]. TIR tracking has been applied to many tasks, including night surveillance and night driver assistants. Although there are many advantages of TIR tracking, it also has some inherent limitations, such as the absence of visual color patterns and low resolutions. Thus, most conventional handcrafted features, e.g. Histograms of Oriented Gradients (HOG) [9] and Color Names [10], cannot represent the TIR object efficiently [11].

Recently, considering the outstanding representation ability of convolutional neural networks (CNN), the deep features which extracted from pre-trained networks are introduced into visual object tracking and obtain more superior tracking performance. These deep features are more discriminative than the convolutional handcrafted features. Last year, by combining the deep motion features which extracted from a pre-trained optical flow network [12] with the standard deep appearance features, DMSRDCF [13] has shown competitive tracking performance. Motivated by these, to enhance the diversity of the limited TIR object features, we exploit both the deep appearance features [6] and deep motion features [13] for TIR tracking in our work.

Most existing tracking approaches exploits the tracking-by-detection paradigm [2], [18]–[21] which treats the tracking problem as a detection task, and trains a regressor or classifier on the object representations. One popular tracking framework is structured output support vector machines (SOSVM). However, the tracking speed of these algorithms is constrained since the high computational complexity [1], [19]. Conversely, training and detecting with cyclic shifts of the training samples and high-dimensional features by solving a ridge regression problem efficiently in Fourier frequency domain [2], [4], discriminative correlation filter (DCF)-based approaches can track the target at high speed for real-time applications.

In this paper, we propose a large margin structured convolution operator to achieve efficient TIR object tracking. By incorporating the DCF and SOSVM, our approach not only inherits the advantage of strong discriminative capability of SOSVM [1], [3], [22] but also obtains surprisingly good results with higher-dimensional features and more dense samples.

The main contributions of our work can be summarized as follows:

- We propose an efficient TIR object tracking algorithm by incorporating the advantages of DCF and SOSVM to employ higher-dimensional features and more dense training samples during tracking.
- We transfer the pre-trained CNNs of visible object tracking to extract deep appearance features and deep motion features of the TIR targets. To the best of our knowledge, we are the first to incorporate the advantages of DCF and SOSVM with deep appearance and motion features for TIR object tracking.
- We suggest an online collaborative optimization strategy to efficiently update the operators, which can improve both tracking performance and speed simultaneously.

Finally, comprehensive experiments are conducted on two modern TIR tracking benchmarks [17], [23]. The results demonstrate that our tracker achieves superior performance both in accuracy and robustness. Comparison results, as shown in Fig.1, illustrate that the proposed tracking approach outperforms several state-of-the-art trackers.

The rest of the paper is organized as follows. We briefly introduce most related works in Section II and describe our proposed tracking algorithm in Section III. Section IV demonstrates the experimental settings and results. Finally, we draw a short conclusion of our work in Section V.

## II. Related Work

In the past decades, SOSVM-based and DCF-based tracking methods have demonstrated favorable performance on numerous object tracking benchmarks [11], [24]. It casts object tracking as a structured output prediction, which admits a consistent target representation for both training and detection. Struck [1] is the first approach exploits SOSVM in tracking, and achieves the best performance on the original OTB benchmark [24]. But the slow computation constraints its extending to high-dimensional features. Recently, LMCF [22] exploits DCF to accelerate SOSVM-based tracking method. Despite SVM-based trackers [3], [22], [25] have provided impressive performances, exploiting high-dimensional features and dense training samples for real-time tracking applications is still a difficult problem.

DCF-based tracking approaches employ the circulant matrices to learn the correlation filters to distinguish the target from its background. The MOSSE tracker [2] first considers training a single-channel correlation filter based on grayscale samples of the target and background appearance. The kernelized correlation filter-based tracker KCF [4] improves MOSSE by introducing kernel methods and multi-dimensional handcrafted features to alleviate the limited ability of linear classification and grayscale features. The recent advancement in DCF based tracking performance is driven by the use of multi-dimensional features [4], [26], adaptively scale estimation [5], [27], reducing boundary effects [6] and multi-resolution features [7], [28].

Due to the impressive capability of feature representations, the combinations of deep CNNs and DCF for visual object tracking have shown state-of-the-art results both in terms of accuracy and robustness. Last year, DMSRDCF [13] have shown excellent tracking performance by fusing all the handcrafted features, deep appearance features and deep motion features. Moreover, MCFT [29] observes that the CNNs which trained on visible images also can well represent the targets for thermal infrared tracking.

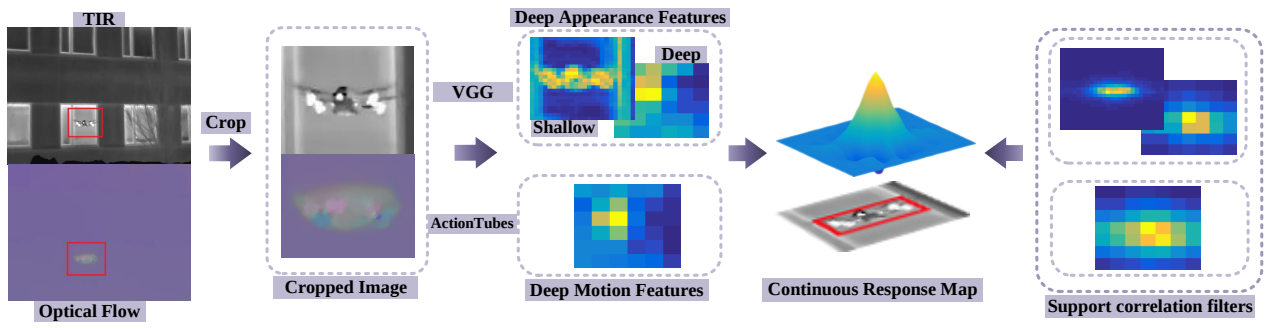

Fig. 2: The framework of our proposed approach.

All the above-mentioned tracking approaches motivate us to investigate the incorporation of DCF and SOSVM frameworks, and transfer the pre-trained appearances network [30] and optical flow network [12] for thermal infrared tracking.

## III. The Proposed Framework

In this section, we first give the overall framework of our large margin structured convolution operator (LMSCO) for TIR object tracking, and then present how to learn the prediction function and update the structured correlation filters by an efficient online collaborative optimization method. The framework of our LMSCO tracker as shown in Fig.2.

### A. Algorithmic overview

We aim to learn a SOSVM-based classifier to distinguish a TIR target from its local backgrounds directly. We consider the TIR tracking problem as a similarity learning task and use a prediction function $f : \mathcal{X} \rightarrow \mathcal{Y}$ based on the input-output pairs to directly estimate the TIR object translational states in a video, where $\mathcal{X}$ denotes the input space and $\mathcal{Y}$ is an arbitrary output space. All the possible cyclic translations of the target centered around the previous position are considered as the training samples, i.e. $\mathcal{Y} = \{\mathbf{y}_p \,|\, p \in \{0, \ldots, M \times N - 1\}\}$, where $M$ and $N$ are the width and height of the image patch. Thus, input-output pairs can be defined as $(\mathbf{x}, \mathbf{y}_p)$, where $\mathbf{x} \in \mathcal{X}$ denotes the target image, and $\mathbf{y}_p \in \mathcal{Y}$ represents one possible observation of $\mathbf{x}$. With different cyclic transform outputs $\mathbf{y}_p$, the pairs stand for different target images which contain diverse translated versions. The joint feature maps of the training samples can be denoted as $\varphi(\mathbf{x}, \mathbf{y}_p)$. We learn a SOSVM with DCF defined by

$$f_{\mathbf{w}}(\mathbf{x}; \mathbf{w}) = \arg\max_{\mathbf{y} \in \mathcal{Y}} \jmath_{\mathbf{w}}(\mathbf{x}, \mathbf{y}) \tag{1}$$

where $\jmath_{\mathbf{w}}(\mathbf{x}, \mathbf{y}) = \langle \mathbf{w}, \varphi(\mathbf{x}, \mathbf{y}) \rangle$ is the confidence score map and the structured correlation filter $\mathbf{w}$ can be learned from the sample set of $\{(\mathbf{x},\mathbf{y}_0),(\mathbf{x},\mathbf{y}_1),\ldots,(\mathbf{x},\mathbf{y}_{M\times N-1})\}$ by solving the following optimization problem:

$$\min_{\mathbf{w}} \|\mathbf{w}\|^2 + C\sum_{p=0}^{MN-1}\xi_p$$
$$\text{s.t. } \forall p:\ \xi_p \geq \ell\{(\mathbf{x},\mathbf{y}_0),(\mathbf{x},\mathbf{y}_p)\} - \{\jmath_{\mathbf{w}}(\mathbf{x},\mathbf{y}_0) - \jmath_{\mathbf{w}}(\mathbf{x},\mathbf{y}_p)\} \tag{2}$$

where $\mathbf{y}_0$ denotes the last predicted position and the slack variables $\xi_p$ represent the penalty assigned to each sample for the margin violations. For desired classified samples, $\xi_p$ will be 0. The regularization parameter $C$ penalizes complicated functions, which are prone to over-fitting, and biases (2) towards training error minimization and margin maximization. $\ell\{(\mathbf{x},\mathbf{y}_0),(\mathbf{x},\mathbf{y}_p)\}$ is a loss function which denotes the structured error associated with a estimated output $\mathbf{y}_p$ when the correct output is $\mathbf{y}_0$. We define the loss function as

$$\ell\{(\mathbf{x},\mathbf{y}_0),(\mathbf{x},\mathbf{y}_p)\} = m_{\mathbf{x}}(\mathbf{y}_0) - m_{\mathbf{x}}(\mathbf{y}_p), \tag{3}$$

where $m_{\mathbf{x}}(\cdot)$ denotes the desired confidence function of the confidence score map $\jmath_{\mathbf{w}}(\mathbf{x},\mathbf{y})$. Here, $m_{\mathbf{x}}(\cdot)$ is designed to follow a Gaussian function as $m_{\mathbf{x}}(\mathbf{y}_p) = \exp(-\frac{(\mathbf{y}_p-\mathbf{y}_0)^2}{2\sigma^2})$ that takes a peak value 1 at the center $\mathbf{y}_0$ and smoothly reduces to 0 for larger cyclic translations.

In order to improve the performance significantly, we follow the recent successful visible object tracking algorithm, namely, C-COT [7], to employ the implicit interpolation model of the training samples in the continuous spatial domain. We consider a $L$ layers feature map to represent the TIR target. In the formulation, we denote feature channel $l \in \{1,2,\ldots,L\}$ of the above-mentioned joint feature map $\varphi(\mathbf{x},\mathbf{y})$, which extracted from the training sample, by $\varphi^l(\mathbf{x},\mathbf{y})$, and each feature layer has an independent resolution $N_l$. Then the feature layer $\varphi^l(\mathbf{x},\mathbf{y}) \in \mathbb{R}^{N_l}$ is viewed as a function $\varphi^l(\mathbf{x},\mathbf{y})[n]$ indexed by the discrete spatial variable $n \in \{0,1,\ldots,N_l-1\}$. Consequently, we can transfer the feature map from the discrete spatial domain to a continuous interval $t \in [0,T)$ by employing an interpolation model $\Phi$ as

$$\Phi\{\varphi^l(\mathbf{x},\mathbf{y})\}(t) = \sum_{n=0}^{N_l-1}\varphi^l(\mathbf{x},\mathbf{y})[n]b_l(t-\frac{T}{N_l}p) \tag{4}$$

where $T$ denotes the size of the continuous spatial domain, $b_l$ represents the interpolation kernel with period $T>0$. In our framework, we employ the cubic spline kernel to construct the interpolation function as described by C-COT [7]. Thus, we can use the continuous $T$-periodic function $\Phi\{\varphi(\mathbf{x},\mathbf{y})\}$ as the entire interpolated joint feature map to the corresponding input-output pairs $(\mathbf{x},\mathbf{y})$.

Therefore, we can also transfer the confidence score map $\jmath_{\mathbf{w}}(\mathbf{x},\mathbf{y})$ to the continuous spatial domain by exploiting the interpolated joint feature map $\Phi\{\varphi(\mathbf{x},\mathbf{y})\}$ of the target image with a set of continuous $T$-periodic multi-layer support convolution filter $\mathbf{w} = (\mathbf{w}^1,\mathbf{w}^2,\ldots,\mathbf{w}^l)$ as

$$\jmath_{\mathbf{w}}(\mathbf{x},\mathbf{y}) = \sum_{l=1}^{L}\mathbf{w}^l \otimes \Phi\{\varphi^l(\mathbf{x},\mathbf{y})\} \tag{5}$$

where $\mathbf{w}^l$ is the continuous support convolution filter for feature layer $l$. The symbol $\otimes$ denotes the circulant convolution operation. Here, to obtain the continuous convolution response, we first interpolate each feature map layer using (4), then convolve the interpolated feature map with its corresponding structured correlation filter. Thus, we can employ the continuous confidence score maps to localize the target in (1) with higher accuracy.

With the continuous interpolation confidence map $S_{\mathbf{w}}(\mathbf{x},\mathbf{y})$, the large margin structured operator can be equivalently formulated as

$$\min_{\mathbf{w}^l} \|\boldsymbol{\gamma}^l\mathbf{w}^l\|^2 + C\sum_{p=0}^{\Lambda-1}\xi_p^l$$
$$\text{s.t. } \forall p:\ \xi_p^l \geq \ell\{(\mathbf{x},\mathbf{y}_0),(\mathbf{x},\mathbf{y}_p)\} - \{\jmath_{\mathbf{w}}^l(\mathbf{x},\mathbf{y}_0) - \jmath_{\mathbf{w}}^l(\mathbf{x},\mathbf{y}_p)\} \tag{6}$$

We add a spatial regularization term in (6) as well as SRDCF [6], which determined by the penalty function $\boldsymbol{\gamma}(t)$, to mitigate the drawbacks of the periodic assumption and control the spatial extent of the structured correlation filters $\mathbf{w}$. The spatial regularization term is also defined in the continuous spatial domain $[0,T)$. In SRDCF [6], this spatial regularization term can make the regularization weights change smoothly from the target image center to the background and enhance the sparsity of the structured correlation filters $\mathbf{w}$ in the Fourier domain.

### *B. Online collaborative optimization*

To solve the problem online, we define a new parameter $\boldsymbol{\epsilon}^l$, where $\boldsymbol{\xi}^l = \boldsymbol{\epsilon}^l + \ell\{(\mathbf{x},\mathbf{y}_0),(\mathbf{x},\mathbf{y})\} - \{\jmath_{\mathbf{w}}^l(\mathbf{x},\mathbf{y}_0) - \jmath_{\mathbf{w}}^l(\mathbf{x},\mathbf{y})\}$, $\boldsymbol{\epsilon} \geq 0$, thus the minimization of (6) is equivalent to the following problem:

$$\min_{\mathbf{w}^l}(\|\boldsymbol{\gamma}^l\mathbf{w}^l\|^2 +$$
$$C\|\boldsymbol{\epsilon}^l + \ell\{(\mathbf{x},\mathbf{y}_0),(\mathbf{x},\mathbf{y})\} - \{\jmath_{\mathbf{w}}^l(\mathbf{x},\mathbf{y}_0) - \jmath_{\mathbf{w}}^l(\mathbf{x},\mathbf{y})\}\|_2^2) \tag{7}$$
$$\text{s.t. } \boldsymbol{\epsilon}^l \geq 0$$

In this formulation, there are two variables $\boldsymbol{\epsilon}^l$ and $\mathbf{w}^l$ to be solved. We observe that when $\mathbf{w}$ is known the subproblem on $\boldsymbol{\epsilon}$ has a closed-form solution, but the subproblem on $\mathbf{w}$ does not have a closed-form solution when $\boldsymbol{\epsilon}$ is known since we exploit the spatially regularization and implicit interpolation components. We employ the Conjugate Gradient method to iteratively solve the subproblem on $\mathbf{w}$. Motivated by SCF [25], we propose an online collaborative optimization method to solve this problem fast and efficiently by iterating between the following two steps.

**Update $\boldsymbol{\epsilon}$.** Given $\mathbf{w}^l$, the subproblem on $\boldsymbol{\epsilon}^l$ becomes

$$\min_{\boldsymbol{\epsilon}^l}\|\boldsymbol{\epsilon}^l + \ell\{(\mathbf{x},\mathbf{y}_0),(\mathbf{x},\mathbf{y})\} - \{\jmath_{\mathbf{w}}^l(\mathbf{x},\mathbf{y}_0) - \jmath_{\mathbf{w}}^l(\mathbf{x},\mathbf{y})\}\|_2^2$$
$$\text{s.t. } \boldsymbol{\epsilon}^l \geq 0 \tag{8}$$

Then the $\boldsymbol{\epsilon}$ subproblem has a closed-form solution

$$\boldsymbol{\epsilon}^l = \max\{0, (\jmath_{\mathbf{w}}^l(\mathbf{x},\mathbf{y}_0) - \jmath_{\mathbf{w}}^l(\mathbf{x},\mathbf{y}) - \ell\{(\mathbf{x},\mathbf{y}_0),(\mathbf{x},\mathbf{y})\})\} \tag{9}$$

**Update w.** Given $\boldsymbol{\epsilon}^l$, the subproblem on $\mathbf{w}^l$ becomes

$$\min_{\mathbf{w}^l}(\|\boldsymbol{\gamma}^l\mathbf{w}^l\|^2 + C\|\jmath^l_{\mathbf{w}}(\mathbf{x},\mathbf{y}_0) - (\jmath^l_{\mathbf{w}}(\mathbf{x},\mathbf{y}) - \ell\{(\mathbf{x},\mathbf{y}_0),(\mathbf{x},\mathbf{y})\} - \boldsymbol{\epsilon}^l)\|_2^2) \tag{10}$$

To obtain a simple expression of the normal equations, we define $\boldsymbol{\rho}^l = \jmath^l_{\mathbf{w}}(\mathbf{x},\mathbf{y}_0) - \ell\{(\mathbf{x},\mathbf{y}_0),(\mathbf{x},\mathbf{y})\} - \boldsymbol{\epsilon}^l$ as the confidence label for each training sample in a continuous spatial domain. Thus, we can rewritten the (10) in Fourier domain as

$$\min_{\mathbf{w}^l} \|\hat{\boldsymbol{\gamma}}^l \otimes \hat{\mathbf{w}}^l\|_2^2 + C\|\hat{\jmath}^l_{\mathbf{w}}(\mathbf{x},\mathbf{y}) - \hat{\boldsymbol{\rho}}^l\|_2^2 \tag{11}$$

where the hat symbol $\wedge$ of a $T$-periodic function represents the Fourier series coefficients, i.e. $\hat{\mathbf{w}}^l[k] = \frac{1}{T}\int_0^T \mathbf{w}^l(t)e^{-i\frac{2\pi k}{T}t}$, $\hat{\boldsymbol{\gamma}}^l[k] = \frac{1}{T}\int_0^T \boldsymbol{\gamma}^l(t)e^{-i\frac{2\pi k}{T}t}$ and $\hat{b}_l[k] = \frac{1}{N_l}\exp(-i\frac{\pi}{N_l})\hat{b}(\frac{k}{N_l})$. Thus the Fourier coefficients of Eq. 5 are obtained as $\hat{\jmath}^l_{\mathbf{w}}(\mathbf{x},\mathbf{y})[k] = \hat{\mathbf{w}}^l[k]\hat{\Phi}^l\{\varphi(\mathbf{x},\mathbf{y})\}[k]\hat{b}_l[k]$, where $\hat{\Phi}^l\{\varphi(\mathbf{x},\mathbf{y})\}[k] = \sum_{n=0}^{N_l-1}\Phi^l\{\varphi(\mathbf{x},\mathbf{y})\}[n]e^{-i\frac{2\pi k}{N_l}}$ is a periodically extended DFT of the interpolated feature maps $\Phi^l\{\varphi(\mathbf{x},\mathbf{y})\}$. The Fourier coefficients of the confidence label is $\hat{\boldsymbol{\rho}}^l[k] = \hat{\jmath}^l_{\mathbf{w}}(\mathbf{x},\mathbf{y}_0)[k] - \hat{\ell}\{(\mathbf{x},\mathbf{y}_0),(\mathbf{x},\mathbf{y})\}[k] - \hat{\boldsymbol{\epsilon}}^l[k]$, where $\hat{\ell}\{(\mathbf{x},\mathbf{y}_0),(\mathbf{x},\mathbf{y})\}[k]$ can be straightforward deduced by $\hat{m_{\mathbf{x}}}\{\mathbf{y}\}[k] = \frac{\sqrt{2\pi\sigma^2}}{T}\exp(-2\sigma^2(\frac{\pi k}{T})^2 - i\frac{2\pi k}{T}\mathbf{y}_0)$. Then the subproblem on $\mathbf{w}^l$ can be addressed by solving the normal equation as

$$\{(\hat{\Phi}^l)^H\hat{\Phi}^l + \frac{1}{C}(\hat{\gamma^l})^H\hat{\gamma}^l\}\hat{\mathbf{w}}^l = (\hat{\Phi}^l)^H(\hat{\boldsymbol{\rho}}^l)^H \tag{12}$$

where $\hat{\mathbf{w}}^H$ represents the complex conjugation of a complex matrix $\hat{\mathbf{w}}$. In our framework, we employ the Conjugate Gradient method as mentioned by C-COT [7] to iteratively solve (12) since it is shown to effectively utilize the sparsity structure. We can easily recover the structured correlation filter $\mathbf{w}$ in the continuous domain with the Fourier transform. Comparing with the canonical correlation filter based trackers, our structured model performs the convolution in the continuous spatial domain by exploiting the implicit interpolated multi-resolution feature maps. Moreover, the proposed approach can obtain the confidence scores of the target straightforward by solving a SOSVM-based prediction function, which is more discriminative than the naïve regression models [4], [5].

In order to improve the performance of TIR tracking, we adopt an efficient scale estimation method [31]. Several scaled regions are considered as the candidate target positions. Then the optimal target position is estimated by maximizing all the scaled confidence score maps over the corresponding location.

## IV. EXPERIMENTS

In this section, we conduct the comprehensive experiments of our proposed algorithm on two TIR tracking benchmarks, i.e. VOT-TIR2015 [23] and VOT-TIR2016 [17]. We first give the implementation details of the experiments. Then we investigate the LMSCO tracker with the impact of several variants of the proposed methods. Finally, we compare LMSCO tracker with several state-of-the-art TIR trackers.

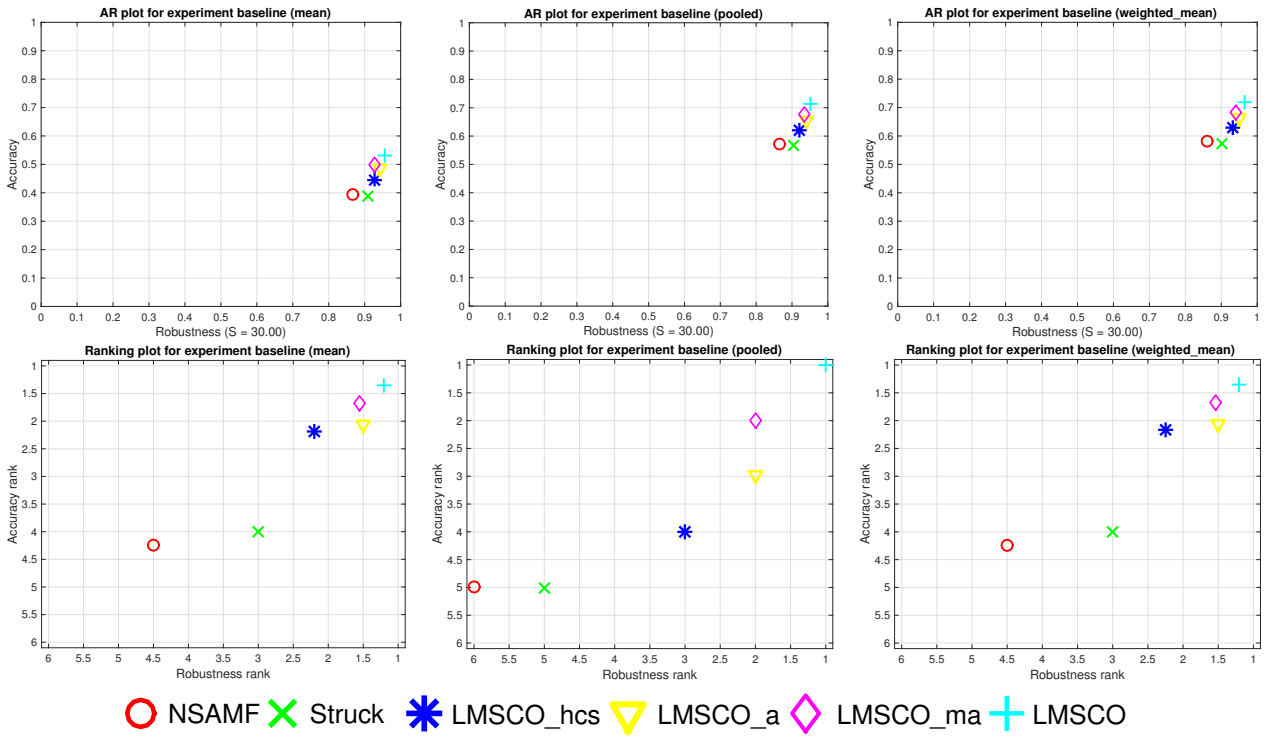

Fig. 3: A-R raw lots and Ranking plots of the experiment baseline with compared trackers on VOT-TIR2015 [23]. The better performance a tracker achieves, the closer it is to the top-right of the plot. Best viewed in color.

### A. *Experimental Settings*

The experiments are conducted in MATLAB R2014b on an Intel(R) Core(TM) i5-4590 CPU @ 3.3GHz with 8GB RAM, and a single NVIDIA GeForce GTX 970 GPU. In our experiments, we crop the original sample image patch to $5^2$ times the previously estimated target bounding box and set the search image patch in a restricted area [200, 300]. As DMSRDCF [13] suggested, we combine the deep appearance features [30], [32] and deep motion features [12] for TIR tracking. The interpolation function $b$ is based on the cubic spline kernel as described in the supplementary material of C-COT [7]. Similar to CFWCR [31], 10 scale layers are used to the adaptive scale estimation. The online collaborative optimization updates in every 5 frames and takes three iterations for each SVM classifiers, while the initial iterations are 30 in the first frame. Specifically, each online collaborative optimization iteration involves two Conjugate Gradient iterations. We keep all parameters fixed in all the following experiments.

### B. *Evaluation Metrics*

There are three evaluation metrics as the VOT challenges committee [11], [33] suggested: (a) the accuracy measurement is obtained from the overlap between the estimated bounding boxes with the ground truth bounding boxes, (b) the robustness measures how many times the overlap area turns out to be zero during the tracking, (c) the expected average overlap (EAO) combines the accuracy and robustness to evaluate the overall performance. The final rank is based on both the accuracy and robustness of each sequence. We use A-R raw plot and A-R ranking plot to visualize the tracking results and EAO score graph to illustrate the overall performance.

### C. *Evaluation on VOT-TIR2015*

In this subsection, we test the effectiveness of each component of our LMSCO method. With different experimental settings, we conduct the experiment on the thermal infrared object tracking benchmark VOT-TIR2015 [23].

**Datasets.** VOT-TIR2015 is the first standard TIR benchmark and has 20 thermal infrared sequences. The benchmark has 6 local attributes, including camera motion, occlusion, object motion, dynamics change and object size change. These local attributes are annotated per-frame in all sequences and the tracking performance of the participating trackers on these attributes can also be compared in the evaluations.

**Compared trackers.** To evaluate the proposed tracking method, we obtain the following four variants of our tracker, which are respectively named as LMSCO-hcs, LMSCO-a, LMSCO-ma and LMSCO-mas. Here, LMSCO-hcs exploits the conventional handcraft features, including HOG [4], [9] and Color Names [10], [26], and uses the scale estimation strategy, LMSCO-a uses the deep appearance features without the scale estimation strategy, LMSCO-ma is the model that combine the deep motion features with the LMSCO-a tracker and LMSCO employ the same features as well as LMSCO-ma and uses the scale estimation strategy. To demonstrate the efficiency of our proposed approach, we also add NSAMF [27] and Struck [1] in our comparison, which are the DCF-based and SOSVM-based tracking approaches using handcrafted features.

**Evaluation and Results.** The results of the participating trackers on the VOT-TIR2015 benchmark are visualized in Figure 3. It is easy to see that LMSCO-hcs achieves better performance than NSAMF and Struck in term of robustness and accuracy. This means that the proposed large margin structured operator is more effective and discriminative. Specifically, LMSCO-ma provides higher robustness and accuracy than LMSCO-a and LMSCO-hcs, which shows that the proposed method is reliable and the deep features, which are extracted from the visible image trained convolutional neural networks, can present the thermal infrared target effectively. Moreover, by comparing LMSCO with LMSCO-ma, we show that the scale estimation strategy can also improve the tracking performance significantly.

### D. *Evaluation on VOT-TIR2016*

In this subsection, we present the evaluation results on the thermal infrared object tracking benchmark VOT-TIR2016 [17] to illustrate our proposed method achieves the impressive results against most state-of-the-art trackers. We employ accuracy, robustness and EAO score as the evaluation metrics to conduct the experiments.

**Datasets.** The thermal infrared tracking benchmark VOT-TIR2016 is an enhanced version of VOT-TIR2015 and has 25 thermal infrared sequences. Compared to the VOT-TIR2015 benchmark, a significant general improvement of results partly compensates for several more difficult sequences. Each sequence also has six local attributes as well as the VOT-TIR2015 which can be used to test the performance of the tracking algorithm on frames with specific attribute evaluation.

**Compared trackers.** For more comprehensive evaluations, we compare our LMSCO with nine state-of-the-art trackers on VOT-TIR2016 benchmark, including MDNet [16], deepMKCF [34], SHCT [15], DSST [5], NSAMF [27], S-RDCF [6], Staple [14], FCT [17] and GGT2 [35]. All these trackers achieve promising results on the previous VOT-TIR challenges.

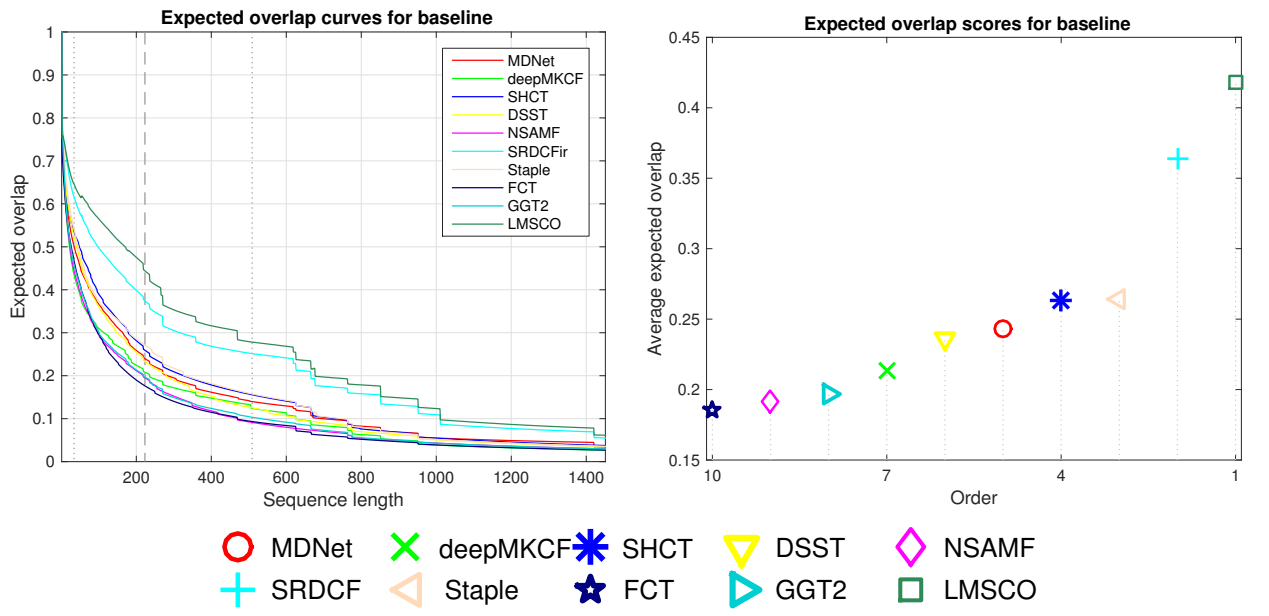

Fig. 4: EAO curve and EAO graph of the experiment baseline with compared trackers on VOT-TIR2016 [17]. In the EAO graph, the better performance a tracker achieves, the closer it is to the right-most of the plot. Best viewed in color.

To ensure a fair and unbiased comparison, we use the original results download from the VOT-TIR2016 challenge homepage (*http://www.votchallenge.net/vot2016/results.html*).

**Evaluation and Results.** Firstly, we evaluate the overall performance of these trackers on the VOT-TIR2016 benchmark. The results are presented in terms of EAO curve and EAO graph as shown in Figure 4. SRDCF is the top-ranked tracker in VOT-TIR2016 and obtains an EAO score of 36.4%. It is clear that our LMSCO achieves the highest EAO score of 42.8% that outperform all the participating trackers and provides an absolute gain of 6.4% in EAO score compared to SRDCF. Then the accuracy and robustness of all these trackers are compared on the VOT-TIR2016 benchmark, the results are shown in Figure 5. Among the compared trackers, LMSCO obtains the best accuracy and robustness in all plots. Although CNNs are trained based on the visible images, it is evident that the fusion of deep appearance features and deep motion features is more suitable compared against the single type deep feature and the conventional handcraft features for TIR tracking.

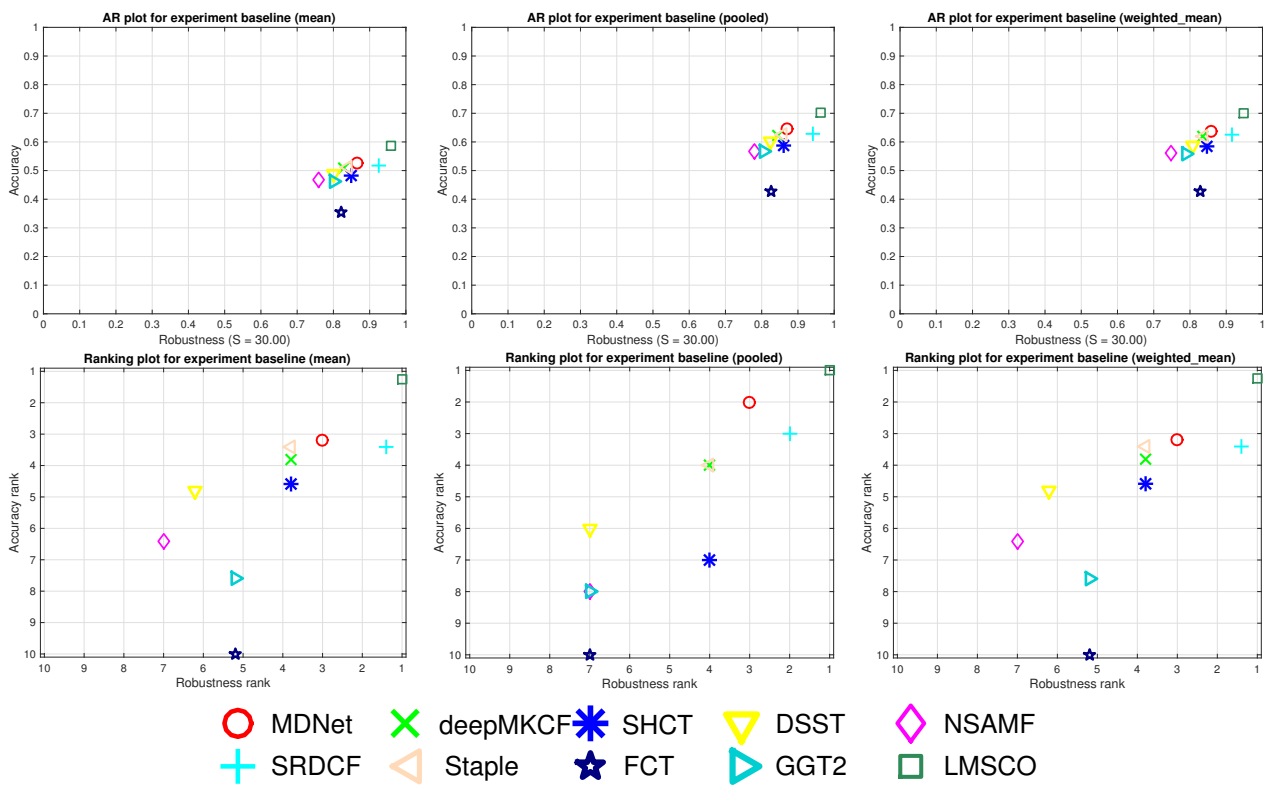

Fig. 5: A-R raw lots and Ranking plots of the experiment baseline with compared trackers on VOT-TIR2016 [17]. The better performance a tracker achieves, the closer it is to the top-right of the plot. Best viewed in color.

## V. Conclusions

In this paper, we propose a TIR object tracking algorithm (LMSCO) based on both the advantages of DCF and SOSVM. We incorporate circulant matrices and correlation filters into the online SOSVM learning algorithm to enable efficient and fast-tracking with higher-dimensional features and more dense training sampling. Then an online collaborative optimization strategy is utilized to update the operator efficiently. Moreover, considering of the powerful representational capability of deep features and their successful applications for visible object tracking, we transfer both the pre-trained CNNs based on visible images and optical flow images for TIR object tracking. Comprehensive evaluations on two modern TIR object tracking benchmarks, i.e. VOT-TIR2015 and VOT-TIR2016, clearly demonstrate that our tracker achieves impressive results and outperforms existing state-of-the-art TIR trackers.

## Acknowledgment

This work is supported by the Science and Technology Planning Project of Guangdong Province, China (Grant No. 2016B090918047).